\documentclass[conference]{IEEEtran}
\IEEEoverridecommandlockouts
\usepackage{cite}
\usepackage{caption}
\usepackage{amsmath,amssymb,amsfonts}
\usepackage{algorithmic}
\usepackage{multicol}
\usepackage{graphicx}
\usepackage{mathtools}
\usepackage{textcomp}
\usepackage{xcolor}
\def\BibTeX{{\rm B\kern-.05em{\sc i\kern-.025em b}\kern-.08em
    T\kern-.1667em\lower.7ex\hbox{E}\kern-.125emX}}
\graphicspath{ {images/} }
    
\begin{document}

\title{Label-Conditioned Next-Frame Video Generation with Neural Flows}

\author{\IEEEauthorblockN{David Donahue}
\IEEEauthorblockA{
\textit{University of Massachusetts Lowell}\\
Lowell, USA \\
david\_donahue@student.uml.edu}}

\maketitle

\begin{abstract}
Recent state-of-the-art video generation systems employ Generative
Adversarial Networks (GANs) or Variational Autoencoders (VAEs) to produce novel videos. However, VAE models typically produce blurry outputs when faced with sub-optimal conditioning of the input, and GANs are known to be unstable for large output sizes. In addition, the output videos of these models are difficult to evaluate, partly because the GAN loss function is not an accurate measure of convergence. In this work, we propose using a state-of-the-art neural flow generator called \textit{Glow} to generate videos conditioned on a textual label, one frame at a time. Neural flow models are more stable than standard GANs, as they only optimize a single cross entropy loss function, which is monotonic and avoids the circular convergence issues of the GAN minimax objective. In addition, we also show how to condition Glow on external context, while still preserving the invertible nature of each "flow" layer. Finally, we evaluate the proposed Glow model by calculating cross entropy on a held-out validation set of videos, in order to compare multiple versions of the proposed model via an ablation study. We show generated videos and discuss future improvements.
\end{abstract}


\section{Introduction}

%Introduction: What is your computer vision research/development question? What is the ultimate goal of your proposed computer vision system? Include image!

%Motivations: Why this is an interesting computer vision question, or why it is important to develop a computer vision system to solve this question?

Text-to-video generation is the process by which a model conditions on text, and produces a video based on that text description. This is the exact opposite of video captioning, which aims to produce a caption that would describe a given video \cite{b6}. It may be argued that text-to-video is a harder task, as there are many more degrees of freedom in pixel space. In addition, video generation is more complex than text-to-image generation \cite{b8}, as a video can be viewed as a collection of images and thus is a superset of the image generation problem. Video generation has many applications, including the automatic generation of animation in educational settings, the realistic generation of sprite movement in video games, as well as other auto-generated entertainment. Reliable video frame generation could play a significant role in the media of tomorrow.

Recently, Generative Adversarial Networks (GANs) have demonstrated promising performance in the domain of real-valued output generation \cite{b7}. Some works are beginning to apply this generative architecture to conditional video generation, with promising results \cite{b1}. However, these GAN systems typically produce all frames of a video at once, in order to ensure coherence across all frames and sufficient error propagation between frames. This is difficult because all video frames, as well as all the intermediate hidden activations needed to produce them must be kept in memory until the final backpropagation step.

\begin{figure*}[ht]
\begin{center}
% \fbox{
\includegraphics[scale=.4]{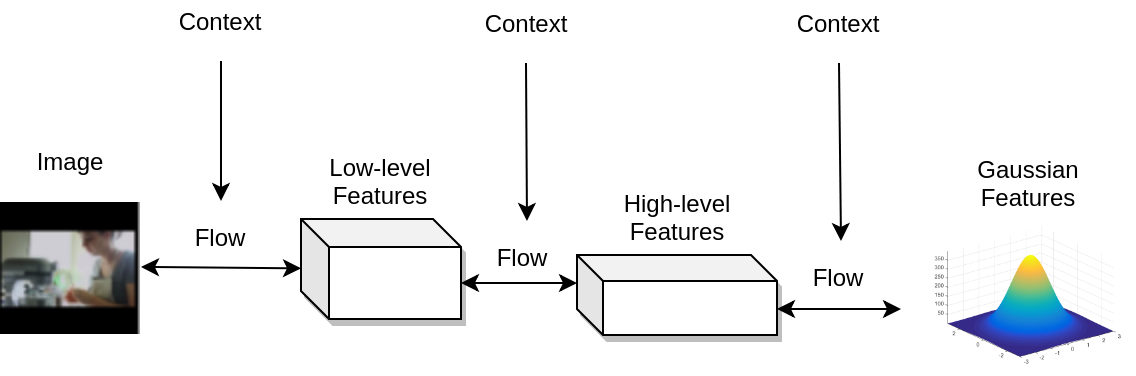}
% }
\caption{Proposed Glow video generation model. Converts dataset image samples into points in Gaussian space, then maximizes the probability of images in that space. During inference, a Gaussian point is sampled and converted back to a realistic image. In this work, we replace images with video frames and add context dependence on previous frames in the video.}
\end{center}
\label{fig:model}
\end{figure*}

In this work, we explore the use of neural flows as a method of video generation in order to solve the following issues:
\linebreak
\begin{enumerate}
  \item Existing Generative Adversarial Networks must generate the entire video at once. We propose to alleviate this problem by introducing a neural flow trained to maximize cross entropy. This model will generate only the next frame, conditioned on all previous frames. This only requires a context encoder over previous frames, with significantly less memory usage
  \item Existing models are difficult to evaluate, as currently GAN-generated instances must be judged by their quality (qualitative). Neural flows instead allow for direct estimation of the probability density of samples, allowing for a quantitative evaluation scheme by measuring the log-probability of examples in a held-out validation set
  \item Generative Adversarial Networks as well as variational autoencoders are unstable without heavy hyper-parameter tuning. We demonstrate the stability of neural flow architectures over previous methods, and in turn show examples of instability and how to improve them
\end{enumerate}
\bigskip

The outline of this paper is as follows \footnote{Computer vision class project}. We introduce recent work in video generation, followed by a description of the system architecture to be implemented. Then we include quantitative cross entropy evaluation of the system, and conduct an ablation study to conclude the effectiveness of each proposed feature. Finally, we support our results with visual frames of generated videos as well as plotted internal model values to justify model behavior. We conclude with future improvements and necessary next steps. Our contributions are as follows:
\linebreak
\begin{itemize}
    \item We utilize the existing Glow neural flow architecture for video generation
    \item We demonstrate how to condition Glow on label features and on a previous frame state
    \item We perform an ablation study of improvements to the architecture
    \item We propose potential future improvements to help convergence and sample quality
\end{itemize}

\section{Recent Work}

%Related work: What kind of existing computer vision research/development work has tried to answer the same or a similar question, what is still unknown? a literature survey (grads)

Compared to other fields of computer vision, such as object classification and detection \cite{b19}, image generation has seen slower progress in comparison. An older choice for video generation was the variational autoencoder (VAE), which utilized an autoencoder with a KL loss component to learn and sample from a continuous latent space. Training points in the latent space are tasked with maximizing a lower bound of the generated data \cite{b15}. This KL loss compactifies the latent space by forcing all training points toward the origin. During inference, a point can be sampled from a Gaussian centered on the origin and decoded to a new training point. However, the compactification of the latent space introduces a loss of information that produces blurry reconstructions. As such, the VAE struggles with the generation of high-quality samples.

Recently, Generative Adversarial Networks have made great strides in a variety of areas \cite{b7}. They work by instantiating two models: a generator network and a discriminator network. The generator attempts to produce samples which are then judged by the discriminator as being real (from the dataset) or fake (from the generator). Error propagation from the discriminator is then used to train the generator to produce more realistic samples. Most work on GANs focus on image synthesis, but GANs can theoretically generate any continuous-valued output \cite{b10}. In the context of video generation, the generator would produce video frames, which the discriminator would then take as input and judge as real or fake. GAN networks can also be conditioned on input, as in the case of text-to-image generation \cite{b8}.

A generative architecture which has received less attention when compared to the Generative Adversarial Network is the neural flow \cite{b16}. Referred to in the paper as RealNVP, this neural flow allows for exact estimation of the probability density of image examples, and allows for direct generation of new images from a Gaussian latent space. This model is trained via a cross entropy loss function similar to that of recent language generation models \cite{b17}, potentially providing more stable training when compared to the minimax objective of Generative Adversarial Networks.

More recently, the Glow architecture was proposed as a modification of the RealNVP neural flow. The Glow model introduces several improvements, such as additive coupling layers and 1x1 invertible convolutions to speed up training time and increase model complexity for better generation \cite{b18}. They apply their model to the task of celebrity image generation. While this model introduces a number of features that the GAN does not possess, such as exact estimation of sample densities, it enforces strict constraints that layers inside the model (flow layers) be invertible and have easily calculable log-determinants. Still, the Glow architecture shows promising results in sample quality.

% include description realnvp and glow architectures

\section{The Neural Flow}

Here we give a brief background of neural flows \cite{b16}. For a given dataset of images $x \in D$ assumed to come from some distribution $p(x)$, we initialize a neural flow to approximate that distribution as $q_\theta(x)$, then maximize the log-likelihood of $N$ dataset examples under this model using the cross entropy objective function:

\begin{equation}
    \mathcal{L} = - \dfrac{1}{N}\sum_{i=1}^{N}\log q_\theta(x_i)
\end{equation}

This minimization has been shown to be equivalent to minimizing the KL-divergence between $p(x)$ and $q_\theta(x)$. This is identical to the objective of existing language generation models, with the exception that the image domain is continuous . A neural flow is able to model a continuous distribution by transforming a simple Gaussian distribution into a complex continuous distribution of any kind. A neural flow model approximates $p(x)$ as follows. It is composed of a series of invertible transformations:

\begin{equation}
    x \xLeftrightarrow{\text{$f_1$}} h_1 \xLeftrightarrow{\text{$f_2$}} h_2 \xLeftrightarrow{\text{$f_n$}} z
\end{equation}

\noindent
We refer to the combined transformation as $f_\theta(x)$, and the inverse transformation as $x = f^{-1}_\theta(x)$. We assign a Gaussian distribution to all points in z-space, and calculate the probability $p(x)$ of any image point $x$ as follows. We restrict each invertible flow layer such that the log-determinant is easily calculable, then by the change of variables formula, we calculate the corresponding z as $z = f_\theta(x)$ and log-probability of $x$ as:

\begin{equation}
    \log q_\theta(x) = \log p(z) + \log \det (dz/dx) = \sum_{i=1}^{K} \log \det (h_{i}/h_{i-1})
\end{equation}

This is intuitive, as the determinant calculation of each layer indicates the stretching and contracting of space, and thus the stretching and contracting of the probability density from layer to layer. Once the model has minimized the cross entropy, image points will correspond to Gaussian points in the high-probability region and an image can be generated by sampling a Gaussian point $z \sim \mathcal{N}(0,1)$, and passing it back through the flow architecture via $x = f^{-1}_\theta(x)$.

Within the model, flow layers are organized into blocks. Between each flow layer, ActNorm is used to normalize the output, which is sent through a 1x1 convolution to allow for shuffling of the input channels.

\begin{table*}[ht]
% \small
\centering
\begin{tabular}{ |r|r|r|r|r| } 
 \hline
 Architecture & Cross Entropy First Frame & Cross Entropy Tail Frames \\
 
 \hline
 Glow + init & 6.4241 & 6.4242\\
 Glow + prev\_frame & \textbf{1.0947} & 1.7243\\
 Glow + state & 1.1217 & \textbf{1.6587}\\
 Glow + state + label & 1.1255 & 1.7193\\

 \hline
\end{tabular}
\caption{\footnotesize Cross entropy values on a held-out validation set for first frame generation and tail frame generation (average of all subsequent frames).}
\label{table:eval}
\end{table*}

\begin{figure*}[ht]
\begin{center}
% \fbox{
\includegraphics[scale=.4]{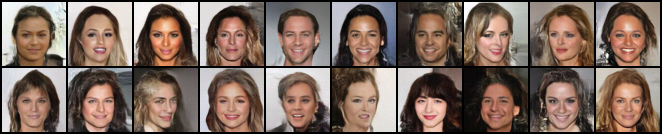}
% }
\caption{The Glow model is able to successfully generate human faces after training on the CelebA dataset.}
\end{center}
\label{fig:faces}
\end{figure*}

\section{Conditional Glow Architecture}

See Figure \ref{fig:model} for a visualization of the Glow architecture. Glow is a type of flow architecture will utilizes a number of unique tricks for generating more high-quality images, while maintaining more flexibility and faster performance. Glow consists of a number of flow layers, each having an identical structure. Each flow consists of an activation normalization layer, an invertible 1x1 convolution, and an affine coupling layer. 

The normalization layer performs similar to batch norm and keeps activations predictible throughout model training. Actnorm works by normalization across the batch dimension only on the first batch of training, then assigns the mean and standard deviation calculations to learnable parameters. As training progresses, the model can learn any values for these parameters. This output passes into the 1x1 convolution, which performs the function of allowing learnable reordering of channels in the input layer. This helps with model performance, due to the invertible nature of each layer. Finally, an affine coupling layer performs a learned transformation by adding a learned transformation of one half of the input to the other half. All layers above have the special property that they are invertible and that the log-determinant is easy to calculate.

We apply the Glow architecture to the problem of video generation. We frame this difficult problem as next-frame generation, and divide total generation into two parts. We initialize a \textit{head model} to produce the first frame of the video, and a \textit{tail model} to generate all subsequent frames. This architecture choice is motivated by the fact that the first frame is the most difficult to generate, and sets up the scene. Subsequent frames work to animate the first frame and continue the plot.

One problem with the proposed tail model is that it must condition on previous frame information in order to generate the next frame. This is a problem, as the original Glow model does not specify a mechanism for conditioning on context information, but rather is utilized for unconditional celebrity image generation \cite{b18}. We augment the Glow model to construct CGlow, to our knowledge the first conditional Glow. As mentioned in the paper, the Glow model is divided into blocks of flow layers, and between each block the input width and height are halved as the number of channels doubles. Thus, a conditioning vector must be resized and provided to all blocks. We achieve this by constructing a pyramid of feature representations from a 2D input, where each convolution-ReLU layer in the pyramid halves the width and height while keeping the channels constant. Each layer of the pyramid is provided to each block of the Glow as input to the additive coupling layer for context.

The tail frame generator could, as a baseline, condition each next frame purely on the previously generated frame. However, this does not capture movement in the video as well as important plot elements. As such we introduce a context processor module, which processes a state vector and the previous frame into a new state vector. The processor consists of an ActNorm layer followed by two convolutions separated by a ReLU activation layer. This processor captures all previous frames into one state, which is provided as context at each tail model generation step, as input to the Glow model using the method outlined above.

\section{Generative Adversarial Network Architecture}

While the final project showcases the use of the Glow architecture outlined above for the purpose of video generation, we also conducted numerous experiments with Generative Adversarial Networks prior to the current work. As a Spring Computer Vision project, work with GANs encompassed approximately two thirds of the entire project, but was scrapped due to poor convergence results. Here we outline the GAN that was constructed, and analyze graphs of its training behavior.

The Generative Adversarial Network consists of a video generator and video discriminator model. Each model employs a modified U-Net architecture \cite{b20} to extract relevant features from the current frame and use them in generation of the next frame. The generator model can be further divided into a first frame generator and tail frame generator, similar to the current work. The U-Net architecture behaves as follows. It receives an input image, and performs a series of convolutional layers separated by ReLU activations. Each convolutional layer halves the width and height dimensions, decreasing the overall image size. This forms a pyramid of image representations, each more abstract than the last. Then, the final representation enters a deconvolutional layer which upsamples to the same size as the second last representation in the constructed pyramid. These are concatenated, and upsampled again to match the size of the following layer. This is repeated until an output representation is produced that is the same size as the input. This layer is taken to be the next frame in the video generation process. 

However, this U-Net architecture can only condition on the previous frame. To enable information to flow across frames, we modify the U-Net to take as input not an image, but a pyramid formed from the previous frame. Each layer of the previous pyramid is conditioned on to produce the pyramid layer for the current frame. To increase information flow and ease of gradients, we introduce skip connections where the output layer of the current pyramid is added to that of the previous frame pyramid via a skip connection. Thus, the pyramid can be seen as changing from frame to frame, and can be interpretted as a multi-layer state passing different levels of information into the future. We refer to this U-Net suited for video generation as View-Net.

To generate the first frame, we utilize a standard U-Net to take a Gaussian tensor of equivalent size to a given frame. This tensor is passed through the U-Net to produce a pyramid of features. These features are then passed into a series of View-Nets to transform the Gaussian tensor into a final pyramid, where the bottom layer is interpreted as the first frame of generation. This frame is passed into N-1 subsequent View-Nets, the bottom layer of each subsequent output pyramid is assumed to be a frame in the video generation process, for a total of N output frames.

The generated frames of the GAN generator are then passed to a discriminator. The discriminator reads the first frame with a standard U-Net, and in a similar process to the generator, utilizes the proposed View-Net to read each subsequent frame. For both generator and discriminator models, parameters are shared between all tail frame generations. The first frame generation layers each have their own separate parameters.

While this model had a number of theoretical benefits, such as easy flow of information and intuitive U-Net next frame generation, it ultimately did not converge. See Figure 7 in the appendix for details. The generated frame mean and standard deviation converged to zero as training progressed, and the generation loss did not decrease with time. In other runs the generation frame output saturated the tanh activation function, producing values as either -1 or +1 but no value in-between. Combined with the observation that the discriminator loss is swiftly minimized, it is possible that the discriminator became too selective at penalizing the generator, such that the only solutions were to maximize or minimize its outputs to avoid exposing variations in the output that the discriminator could use to identify those examples. This demonstrates an important issue with GANs, namely that they can be unstable \cite{b21}. This issue is likely compounded for larger output spaces such as videos, where there are more dimensions of the generator output that the discriminator can use to identify false examples.

Ultimately, the lack of convergence for the GAN model motivated a switch to the Glow architecture, for its increased stability and arguably simpler design, as well as the benefit of cross entropy evaluation of flow models.

\section{Chosen Dataset}

In choosing a video generation dataset, we wanted a simple dataset that could evaluate the core design of the proposed architecture while still providing quality samples. We utilize the Moments in Time dataset consisting of 100,000 videos. Each video is 256x256 and consists of 90 frames for 3 seconds of play time. To improve development time, we downsample to 64x64 and report results after training on the first 30 frames (1 second). The Moments in Time dataset is divided into 201 categories which provide a one word description of the content of the video. For example, the "walking" category contains 500 videos of persons walking through various environments. We split the dataset into train (99\%) and validation (1\%) splits and used a fixed seed to avoid mixing of both datasets during evaluation.

\begin{figure}[t!]
\begin{center}
\includegraphics[scale=.5]{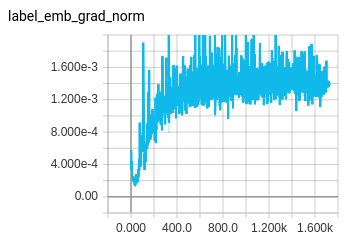}
\caption{Gradient norm of label embeddings.}
\label{fig:lembs}
\end{center}
\end{figure}

\section{Evaluation}

%Evaluation: What computer vision data set(s) do you plan to use to validate your proposed approach? How are you going to evaluate your solution? In another word, what is your plan to demonstrate that your solution/answer is good or is reasonable? If applicable, add any of your initial results.

Popular GAN papers provide visual examples of generated images, but do not provide exact measurements of sample quality in a quantitative way. One major benefit of neural flow models is explicit evaluation - this can be performed by evaluating the log probability of examples from the validation set that the model has not yet seen. If the flow model has successfully captured the distribution in question, then the log probability of samples from that distribution will be maximized. It is important that this evaluation not be done on only the training set, as the model may overfit to these examples and thus not generalize well in capturing the diversity of the data.

We report cross entropy (mean negative log probability) of sampled videos from the validation set in Table 1. This cross entropy is reported per frame in the video. We report head and tail cross entropies separately, to explore the difference in quality between the two models and the difference in task difficult of generating the first and subsequent frames.

As mentioned previously, GAN and VAE models do not have an explicit quality evaluation metric similar to the cross entropy approach of the neural flow, and thus we do not include them as a baseline in this evaluation scheme. Instead, we perform an ablation study, and investigate the effects of adding different components in the paper. First, we show a randomly-initialized Glow model with no conditioning (Glow + init), meaning that each frame is generated independently of all others. Next, we condition the tail frame CGlow generator on only the previous frame (Glow + prev\_frame). We then add the context processor to provide state information in addition to the previous frame (Glow + state). Both the previous frame and the state are concatenated and provided to the CGlow tail generator. Finally, we provide label information to the first frame Glow generator (now CGlow) by learning a 3-channel image with identical size to a video frame, and provide this representation as input to the head generator for context during first frame generation (Glow + state + label). See Table \ref{table:eval} for results on the validation set.

In addition, we show samples generated from the model. To demonstrate that the model can successfully generate images of interest, we train the model on the CelebA dataset and show the resulting faces in Figure 2. Then, we visualize the first frame generator outputs through visual examples in Figure 4. Further, we show generated videos from the model after training on the entire dataset, as well as frames generated after overfitting to a single video with four frames (Figure 5).

\section{Results}

Overall, the results of this analysis were different than expected. Overall, all trained models outperformed the randomly-initialized model variant for both head and tail frame generation. This suggests that the model is able to significantly improve its approximation of the video distribution through time. In addition, providing context in the form of a state vector seemed to improve cross entropy performance by 0.027 points. While this is a small number, cross entropy gains tend to be small in the measured range of 1 to 2 points. Still, reliance on the state vector could be improved, but it helped in overall model performance. One discouraging result is that label information was not properly utilized, as the first frame generator performed worse with the addition of label features.

\section{Discussion}

For this, we used the head first frame generator to synthesize images after training on the CelebA dataset. The generated samples look realistic, and demonstrate that the model can perform well when the variance of the dataset is limited.

From the results of the validation cross entropy analysis, state vector context information seemed to improve performance on the task. This is expected as a model that only has access to the previous frame would face difficulty in determining the direction of motion for the next frame. Context information would serve to provide that directional information for better predictions.

It is surprising that label information was of no use to the model. The label information was provided in the form of a learned tensor, which the model could fill with any helper information available. In addition, label information helps to narrow the scope of generation and indicate to the model exactly what category of video it should generation. This can greatly help convergence. When attempting to diagnose why the model does not exploit label information, it is helpful to analyze the gradients received by the learned label representations themselves. See Figure 3 for details on the label embedding norm. The norm is in the order of $10^-6$, which seems to be extremely low compared to the tail model, which has an equivalent number of parameters. This indicates that the label embeddings are not training, and explains why the model is unable to exploit them for boosted performance on this task. The reason for this lack of training should be explored further.

\begin{figure*}[ht]
\begin{center}
% \fbox{
\includegraphics[scale=.3]{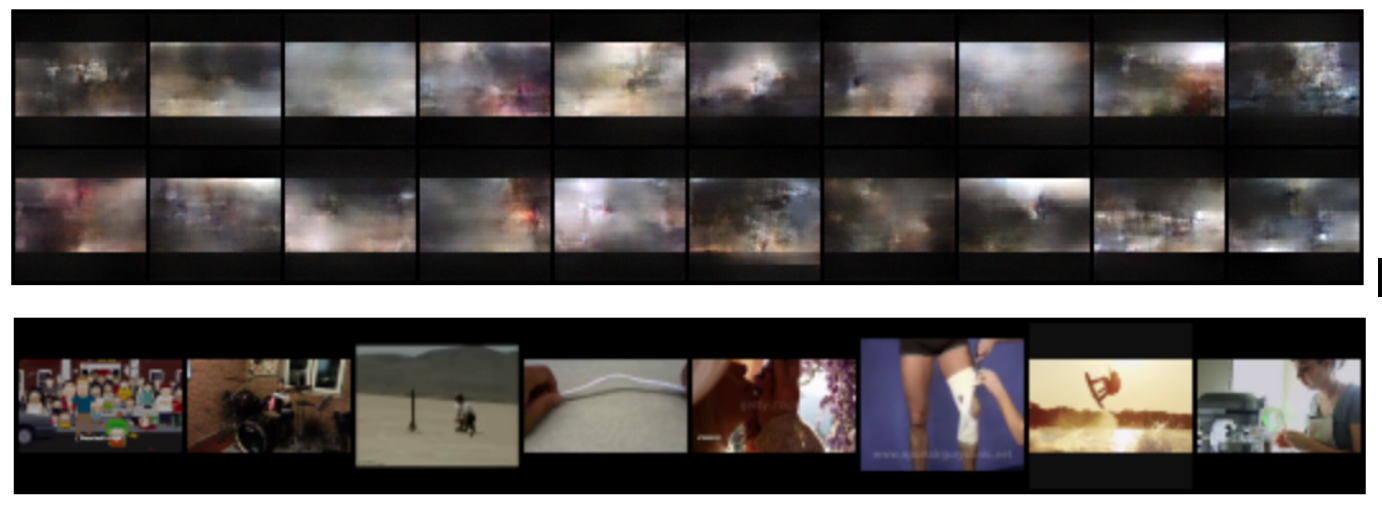}
% }
\caption{Here are samples of the first frame of videos generated by the model (top), and real first frames from the dataset (bottom).}
\end{center}
\label{fig:faces}
\end{figure*}

\begin{figure*}[ht]
\begin{center}
% \fbox{
\includegraphics[scale=.2]{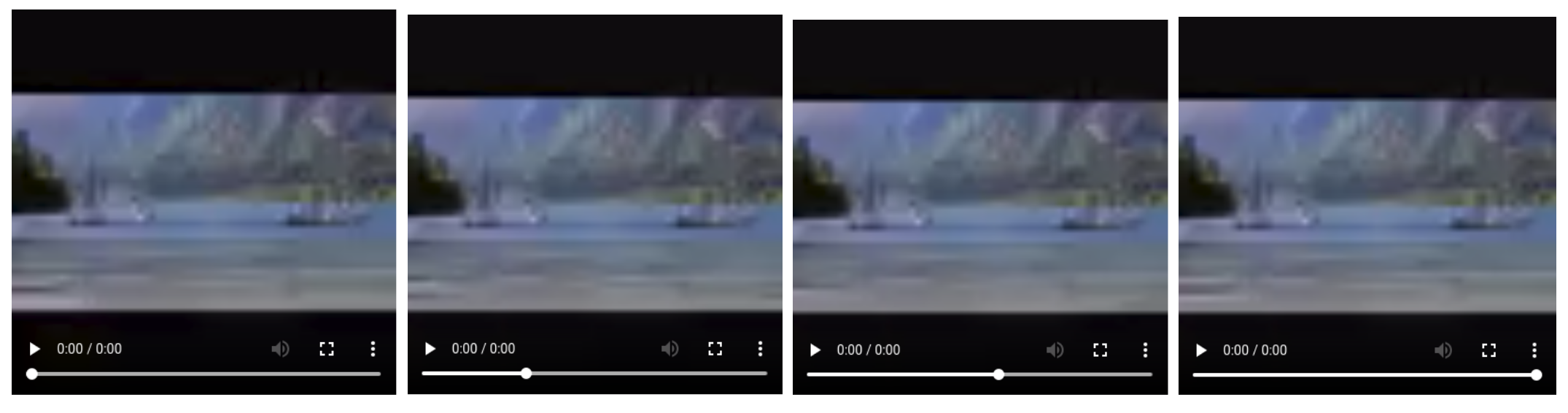}
% }
\caption{Demonstration of the model training on the first four frames of a single video. This demonstrates that the model has the capacity to learn the dynamics of a single video. The boat moves slightly to the right over the video.}
\end{center}
\label{fig:faces}
\end{figure*}

\begin{figure*}[ht]
\begin{center}
% \fbox{
\includegraphics[scale=.15]{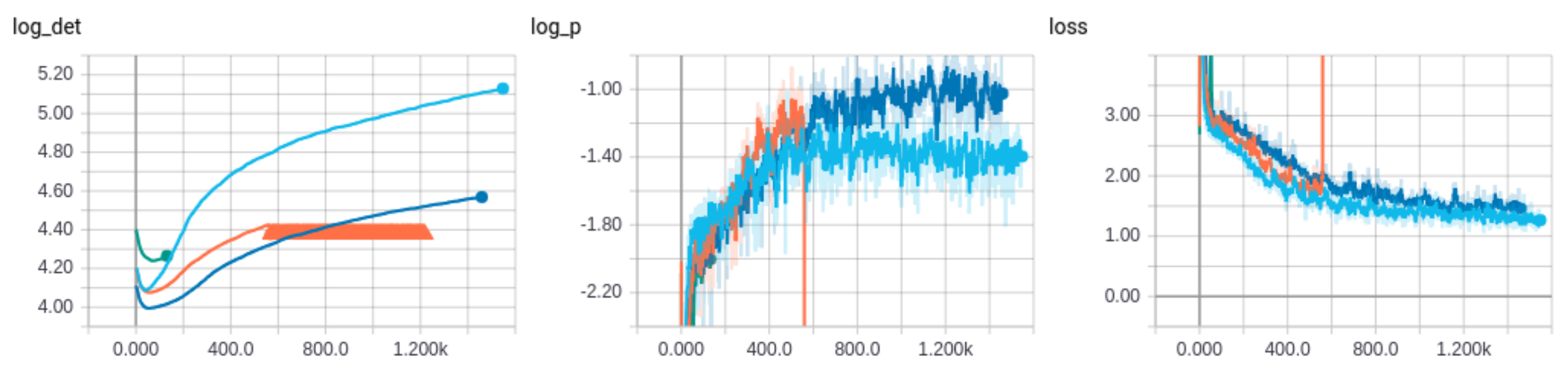}
% }
\caption{Multiple GAN model runs reporting log determinant, log probability of training examples, and cross entropy loss (left), each in a different color. Orange run shows spontaneous burst in loss value followed by NaN (not shown).}
\end{center}
\label{fig:faces}
\end{figure*}

There were many important barriers to overcome during the training and correction of this Glow architecture. First, while the base Glow model performed well on the generation of human faces (see Figure 2), it faced many challenges when transfering to the video domain. The first apparent difference between the datasets is variance. In the CelebA dataset, faces do not differ that widely; all are in a similar orientation and at a similar camera angle. The background of each photo is relatively simple. However, the videos observed in the Moments in Time dataset exhibit significant variety, with over 200 categories of video taken by real people. When comparing Figure 4 and Figure 5, it is possible that the variety of the dataset used to train the model tended to confuse video generation, resulting in blurry images that do not contain any distinct objects. This training setup appears to overwhelm the model. This effect could have been produced because the model did not train for a significant amount of time (5 hours), or due to model capacity (the tail model is relatively small), but this could have also occured due to the training objective. The cross entropy objective penalizes the negative log-probability of examples from the dataset. If the model chose to ignore a specific mode of image in favor of increased sample quality, it would be penalized. In the extreme case, the model might ignore an example by assigning near zero probability; this would result in a loss near $-log(0)$ or near infinity. As such, the model cannot ignore any one example but must capture the entire variety of the dataset at once.

The variance of the dataset may have caused a second issue during model training: spontaneous spiking of the loss value during training and subsequent NaN values which filled all statistics. Figure 6 shows this phenomenon. The loss value spikes suddenly, causing model instability and subsequent NaN values for the remainder of training. As the model becomes more confident and assigns higher probability to examples seen in the training set, it pushes videos that are disimilar to those in the training set to regions of low-probability. In a multivariate Gaussian distribution, these regions are far from the origin and therefore have large magnitudes. When a training example arises that is wildly different from what the model has seen previously, it can land far from the origin, and the resultant magnitude of the output can cause numerical instability. This numerical stability could likely have caused the NaN values observed after just two hours of training. To combat these NaN values, I introduced gradient norm clipping, which truncated the magnitude of gradients if they surpassed as specific value. This kept the model from learning to fast in the presence of low-probability examples, and likely sped up convergence. As a backup, the model would save periodically and load a previous checkpoint in the presence of detected NaN values, but this instability occurred rarely and did not result in a significant loss in training time.

A claim of this paper is that the context processor reduces the computational complexity of generation. Since the context generator only contains approximately 10,000 parameters, it is very lightweight and only undergoes two convolutions per frame. This is small in comparison to the approximately 16 convolutions per each tail frame. This can save memory if previous glow generations are freed from memory.

\section{Conclusion}

In this paper, we demonstrated the use of the Glow generation architecture on the task of video generation. We showed that providing the Glow with a context representation of previous frames aids in better video quality and cross entropy evaluation. In this work, we introduced numerous visualizations of generator outputs and graphs indicating model behavior and challenges faced throughout the project. We visualized the results of a separate GAN model, and demonstrated its instability for the task.

\pagebreak
\appendix

\section{Hyperparameters}

For the first frame Glow generator, we initialize the model with 4 blocks of 32 flow layers each. We represent each image as a 3x64x64 image as input to the model during training. The context processor consists of a state input size of 16, which passes its input through ActNorm, convolves the state + previous frame (19 channels) into double the state size (32), passes the result through a ReLU activation and convolves the output again to match the size of the input state. This result is added to the input state via a skip connection to produce the next state, which is passed to the Glow generator.

Conditioning the Glow generator involved taking an input, and passing it through 4 convolutional downsampling layers, one for each block. The first took the 19x64x64 and downsampled it to 19x32x32, then passed it through a ReLU activation, followed by another downsampling convolution to 19x16x16, followed by another downsampling convolution and ReLU to a final 19x8x8. Each representation was passed as input to each corresponding block, and finally passed as input to the convolutional layer producing the translation vector for the additive coupling layer.

All other Glow parameters match those in the Glow paper \cite{b18}. The model was trained with a learning rate of 1e-4 at a batch size of 8 for one pass over the training dataset (5 hours).

There were 201 video labels/categories in the Moments in Time dataset. When conditioning on label embeddings, an embedding matrix was initialized of size 201x3x64x74. During training, the embeddings for the labels of each video were looked up and provided as input to the first frame generator, which conditioned on this 3 channel input in an identical way to the state vector conditioning of the tail frame generator.

While the model diagram illustrates a simplistic mapping from an image point to a single Gaussian representation, the actual Glow implementation produces multiple Gaussian tensors of decreasing size from the image input to the final layer. The probability of training points is calculated in the entire combined space of all the produced Gaussian representations.

\section{Graphs and Visualizations}

Here we introduce various visualization and graphs utilized to show model performance and training statistics.

\begin{figure*}[ht]
\begin{center}
% \fbox{
\includegraphics[scale=.2]{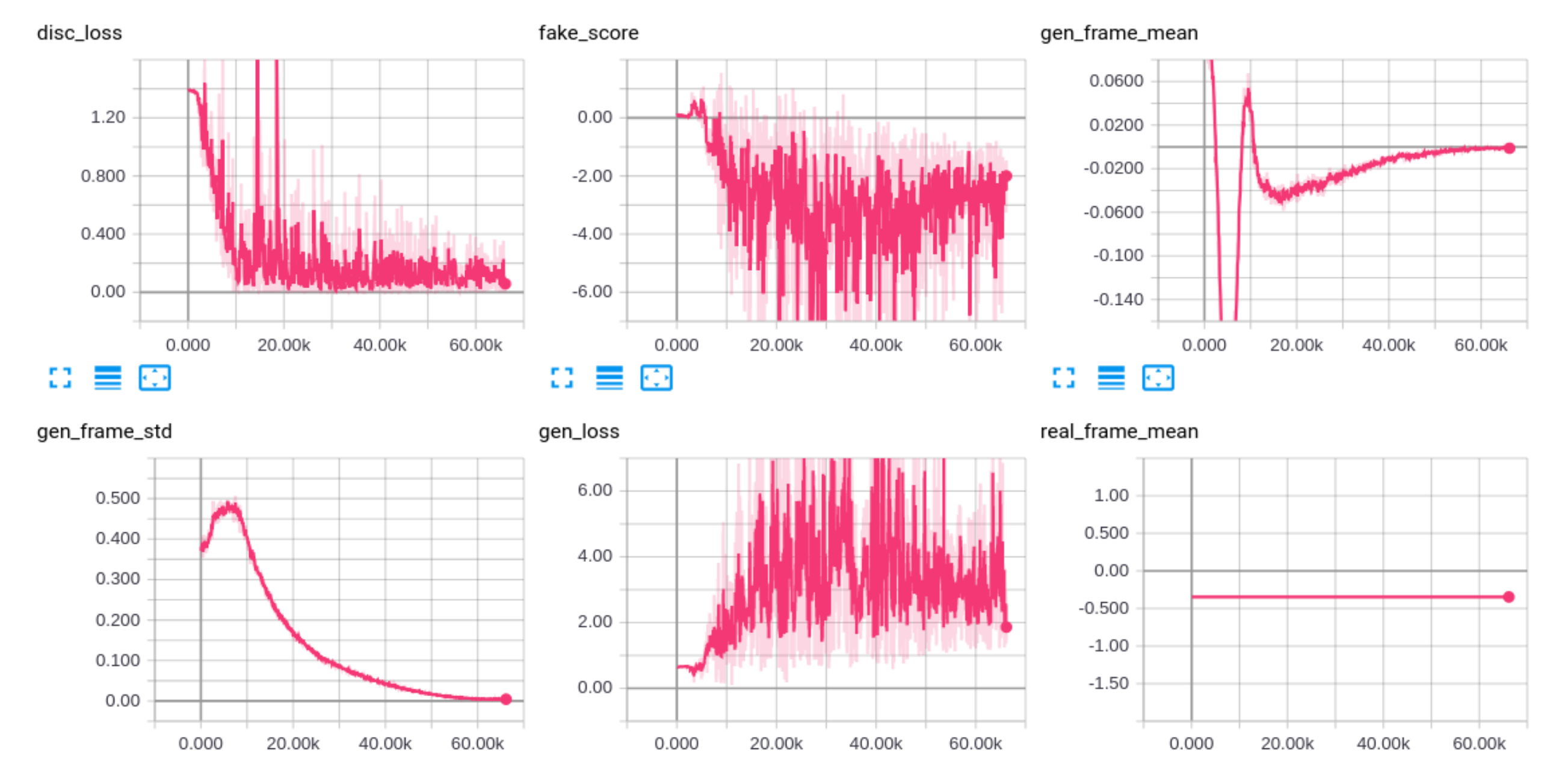}
% }
\caption{Graphs visualizing GAN training. Instabilities in the model caused the mean and standard deviation (top right and bottom left respectively) to converge to zero. In other training setups, these values saturated to the maximum values of the tanh activation (-1 and +1).}
\end{center}
\label{fig:faces}
\end{figure*}

\begin{figure*}[ht]
\begin{center}
% \fbox{
\includegraphics[scale=.2]{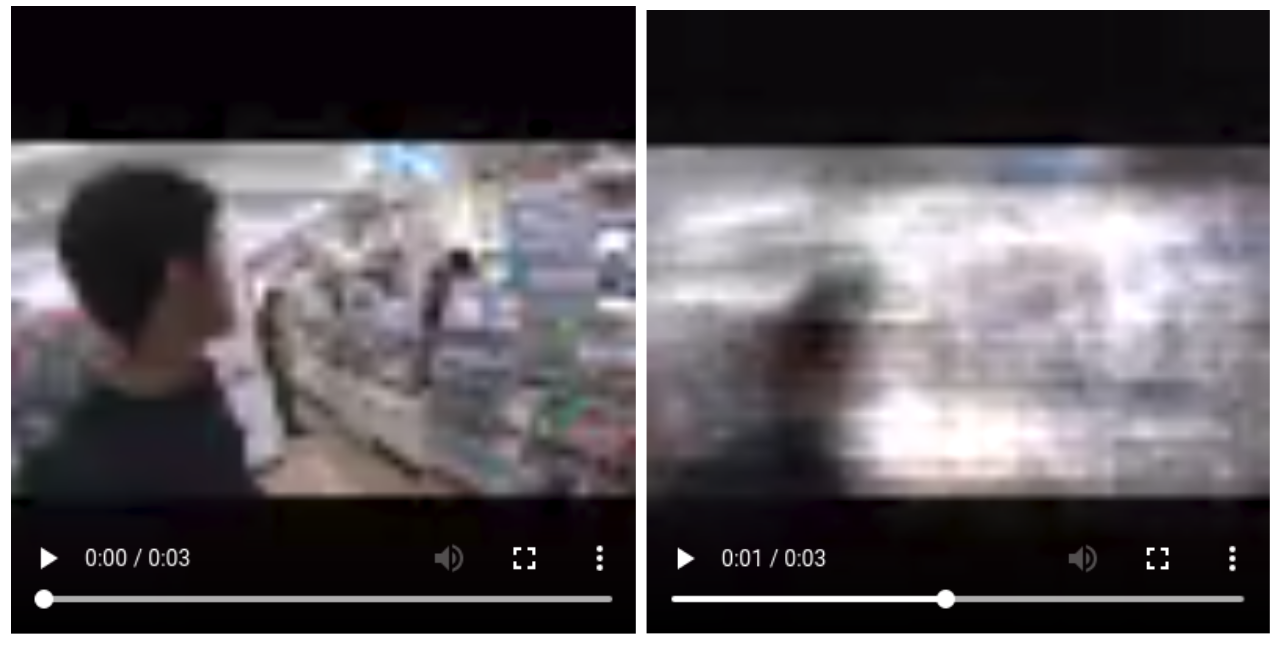}
% }
\caption{This shows the proposed Glow generation model overfitting on a single video. While the first frame generator is powerful enough to model the first frame, the tail generator struggles to maintain coherence throughout the remainder of the video.}
\end{center}
\label{fig:faces}
\end{figure*}

\end{document}